# Expectation Maximization and Complex Duration Distributions for Continuous Time Bayesian Networks

**Uri Nodelman**
Stanford University
nodelman@cs.stanford.edu

**Christian R. Shelton**
UC Riverside
cshelton@cs.ucr.edu

**Daphne Koller**
Stanford University
koller@cs.stanford.edu

## Abstract

Continuous time Bayesian networks (CTBNs) describe structured stochastic processes with finitely many states that evolve over continuous time. A CTBN is a directed (possibly cyclic) dependency graph over a set of variables, each of which represents a finite state continuous time Markov process whose transition model is a function of its parents. We address the problem of learning the parameters and structure of a CTBN from partially observed data. We show how to apply expectation maximization (EM) and structural expectation maximization (SEM) to CTBNs. The availability of the EM algorithm allows us to extend the representation of CTBNs to allow a much richer class of transition durations distributions, known as *phase distributions*. This class is a highly expressive semi-parametric representation, which can approximate any duration distribution arbitrarily closely. This extension to the CTBN framework addresses one of the main limitations of both CTBNs and DBNs — the restriction to exponentially / geometrically distributed duration. We present experimental results on a real data set of people's life spans, showing that our algorithm learns reasonable models — structure and parameters — from partially observed data, and, with the use of phase distributions, achieves better performance than DBNs.

## 1 Introduction

Many applications involve reasoning about a complex system that evolves over time. Standard frameworks, such as hidden Markov models (HMMs) (Rabiner & Juang, 1986) and dynamic Bayesian networks (DBNs) (Dean & Kanazawa, 1989), discretize time at fixed intervals, known as time slices, and then model the system as evolving discretely from one time slice to the next. However, in many systems, there is no natural time granularity: Some variables evolve quickly, whereas others change more slowly; even the same variable can change quickly in some conditions and slowly in others. In such systems, attempts to model the system as evolving over uniform discrete time intervals either lead to very coarse approximations, or require that the entire trajectory be modelled at a very fine granularity, at a high computational cost. This problem is even more acute in a learning setting, when little prior knowledge might exist about the rate of evolution of different variables in the system.

Another approach is to model such systems as evolving over continuous time. For discrete state systems, such representations include event history analysis (Blossfeld et al., 1988; Blossfeld & Rohwer, 1995) and Markov process models (Duffie et al., 1996; Lando, 1998). Nodelman et al. (2002) extend these representations with ideas from the framework of Bayesian networks, to define *continuous time Bayesian networks (CTBNs)* — a structured representation for complex systems evolving over continuous time.

An important task for any model is constructing it. One approach for acquiring models that fit the domain is by learning them from data samples. Nodelman et al. (2003) present an algorithm for learning a CTBN — both structure and parameters — from fully observed data. However, in many real-world applications, we obtain only partial observations of the trajectory. This issue is likely to be even more acute when learning continuous time models where it is difficult to observe everything all the time. Therefore, learning networks from partially observed data is an important problem with real-world significance.

In this paper, we provide an algorithm, based on the Expectation Maximization (EM) algorithm (Dempster et al., 1977), for learning CTBN parameters from partially observable data. We also provide an extension, based on structural EM (Friedman 1997; 1998), for learning the network structure from such data. Our algorithm also provides us with a solution to one of the major limitations of both CTBN and DBN models — their use of a transition model which evolves exponentially (or geometrically) with time. In particular, we build on the rich class of *phase distributions* (Neuts 1975; 1981), showing how to integrate them into a CTBN model, and how to use our EM algorithm to learn them from data. We present results of our learning algorithm on real world data related to people's life histories, and show that our learned CTBN model with phase distribution transitions is a better model of the data than both learned DBN and learned CTBN models.

## 2 EM for Markov Processes

We begin by reviewing the process of EM for homogeneous Markov processes, on which our algorithm for CTBNs is strongly based. Variants of this problem have been addressed by various authors, including Sundberg (1974; 1976), Dembo and Zeitouni (1986), Asmussen et al. (1996), and Holmes and Rubin (2002). Our presentation is based on the formulation of Asmussen et al. (1996).

### 2.1 Homogeneous Markov Processes

A finite state, continuous time, homogeneous Markov process $X_t$ with state space $Val(X) = \{x_1, \ldots, x_n\}$ is described by an initial distribution $P_X^0$ and an $n \times n$ matrix of transition *intensities*:

$$\mathbf{Q}_X = \begin{bmatrix} -q_{x_1} & q_{x_1 x_2} & \cdots & q_{x_1 x_n} \\ q_{x_2 x_1} & -q_{x_2} & \cdots & q_{x_2 x_n} \\ \vdots & \vdots & \ddots & \vdots \\ q_{x_n x_1} & q_{x_n x_2} & \cdots & -q_{x_n} \end{bmatrix},$$

where $q_{x_i x_j}$ is the intensity of transitioning from state $x_i$ to state $x_j$ and $q_{x_i} = \sum_{j \neq i} q_{x_i x_j}$. The intensity $q_{x_i}$ gives the "instantaneous probability" of leaving state $x_i$ and the intensity $q_{x_i x_j}$ gives the 'instantaneous probability' of transitioning from $x_i$ to $x_j$. More formally, as $\Delta t \to 0$,

$$\Pr\{X(t + \Delta t) = x_j \mid X(t) = x_i\} \approx q_{x_i x_j} \Delta t, \text{ for } i \neq j$$
$$\Pr\{X(t + \Delta t) = x_i \mid X(t) = x_i\} \approx 1 - q_{x_i} \Delta t .$$

The transient behavior of $X_t$ is as follows. Variable $X$ stays in state $x$ for time $T$ exponentially distributed with parameter $q_x$. The probability density function $f$ for $X_t$ remaining at $x$ is $f(q_x, t) = q_x \exp(-q_x t)$ for $t \geq 0$, and the expected time of transition is $1/q_x$. Upon transitioning, $X$ shifts to state $x'$ with probability $q_{xx'}/q_x$.

The distribution over transitions of $X$ factors into two pieces — exponential for *when* the next transition occurs and multinomial for *where* the state transitions, i.e., the next state. The natural parameters are $q_x$ for the exponential distribution and $\theta_{xx'} = q_{xx'}/q_x, x' \neq x$ for the multinomial distribution.

The distribution over the state of the process $X$ at some future time $t$, $P_X(t)$, can be computed directly from $\mathbf{Q}_X$. If $P_X^0$ is the distribution over $X$ at time 0, then

$$P_X(t) = P_X^0 \exp(\mathbf{Q}_X t) ,$$

where $\exp$ is matrix exponentiation.

### 2.2 Incomplete Data

For a Markov process $X$, our data are a set of partially observed trajectories $\mathcal{D} = \{\sigma[1], \ldots, \sigma[w]\}$ that describe the behavior of $X$. A complete trajectory can be specified as a sequence of states $x_i$ of $X$, each with an associated duration. This means we observe every transition of the system from one state to the next and the time at which it occurs. In contrast, a partially observed trajectory $\sigma \in \mathcal{D}$ can be specified as a sequence of *subsystems* $S_i$ of $X$, each with an associated duration. A subsystem is simply a nonempty subset of states of $X$, in which we know the system stayed fur the duration of the observation. Some transitions are partially observed — we know only that they take us from one subsystem to another. Transitions from one state to another within the subsystem are wholly unobserved — hence we do not know how many there are nor when they occur.

Note that momentary observation of the state of the system, or *point evidence* can also described in this framework. In particular, in the sequence of observed subsystems, durations can zero.

For each trajectory $\sigma[i]$ we can consider a space $\mathbf{H}[i]$ of possible completions of that trajectory. Each completion $h[i] \in \mathbf{H}[i]$ specifies, for each transition of $\sigma[i]$, which underlying transition of $X$ occurred and also specifies all the entirely unobserved transitions of $X$. Combining $\sigma[i]$ and $h[i]$ gives us a complete trajectory $\sigma^+[i]$ over $X$. Note that, in a partially observed trajectory, the number of possible unobserved transitions is unknown. Moreover, there are uncountably many times at which each transition can take place. Thus, the set of possible completions of a partial trajectory $\sigma$ is, in general, the union of a countably infinite number of spaces, which are real-valued spaces of unbounded dimension. Nevertheless, the notion of all possible completions is well-defined. We can define the set $\mathcal{D}^+ = \{\sigma^+[1], \ldots, \sigma^+[w]\}$ of completions of all of the partial trajectories in $\mathcal{D}$.

**Example 2.1** *Suppose we have a process $X$ whose state space is $Val(X) = \{y_1, y_2\} \times \{z_1, z_2\}$. Consider the following example $\sigma^+$ of a fully observed trajectory over the time interval $[0, 2)$: $X$ starts in $\langle y_1, z_2 \rangle$ at time 0; at time 0.5 it transitions to $\langle y_2, z_2 \rangle$; at time 1.7 it transitions to $\langle y_2, z_1 \rangle$.*

*Note we can write $\langle \cdot, z_i \rangle$ for the subsystem consisting of the states $\langle y_1, z_i \rangle$ and $\langle y_2, z_i \rangle$. An example partially observed trajectory $\sigma$ over the interval $[0, 2)$ is: $X$ starts in $\langle \cdot, z_1 \rangle$; at time 1.7 it transitions to $\langle \cdot, z_2 \rangle$. Note that $\sigma^+$ is a completion of $\sigma$. Another possible completion of $\sigma$ is: $X$ starts in $\langle y_2, z_1 \rangle$ at time 0; at time 1.0 it transitions to $\langle y_1, z_1 \rangle$; at time 1.7 it transitions to $\langle y_1, z_2 \rangle$.*

*We can describe a partially observed trajectory $\sigma'$ with point evidence at time 0.7 and 1.8: $X$ starts in $\langle \cdot, \cdot \rangle$; at time 0.7 we observe $\langle \cdot, z_2 \rangle$; from time 0.7 to 1.8 we observe $X$ in $\langle \cdot, \cdot \rangle$; at time 1.8 we observe $X$ in $\langle \cdot, z_1 \rangle$; from time 1.8 on, we observe $X$ in $\langle \cdot, \cdot \rangle$. Note that $\sigma^+$ is also a completion of $\sigma'$.*

## 2.3 Expected Sufficient Statistics and Likelihood

The sufficient statistics of a set of complete trajectories $\mathcal{D}^+$ for a Markov process are $T[x]$ — the total amount of time that $X = x$, and $M[x, x']$ — the number of times $X$ transitions from $x$ to $x'$. If we let $M[x] = \sum_{x'} M[x, x']$ we can write the log-likelihood for $X$ (Nodelman et al., 2003):

$$\ell_X(\boldsymbol{q}, \boldsymbol{\theta} : \mathcal{D}^+) = \ell_X(\boldsymbol{q} : \mathcal{D}^+) + \ell_X(\boldsymbol{\theta} : \mathcal{D}^+)$$
$$= \sum_x \left( M[x] \ln(q_x) - q_x T[x] + \sum_{x' \neq x} M[x, x'] \ln(\theta_{xx'}) \right).$$

Let $r$ be a probability density over each completion $\boldsymbol{H}[i]$ which, in turn, yields a density over possible completions of the data $\mathcal{D}^+$. We can write the expectations of the sufficient statistics with respect to the probability density over possible completions of the data: $\bar{T}[x]$, $\bar{M}[x, x']$, and $\bar{M}[x]$. These expected sufficient statistics allow us to write the expected log-likelihood for $X$ as

$$\mathbf{E}_r[\ell_X(\boldsymbol{q}, \boldsymbol{\theta} : \mathcal{D}^+)] = \mathbf{E}_r[\ell_X(\boldsymbol{q} : \mathcal{D}^+)] + \mathbf{E}_r[\ell_X(\boldsymbol{\theta} : \mathcal{D}^+)]$$
$$= \sum_x \left( \bar{M}[x] \ln(q_x) - q_x \bar{T}[x] + \sum_{x' \neq x} \bar{M}[x, x'] \ln(\theta_{xx'}) \right).$$

## 2.4 The EM Algorithm

We use the expectation maximization (EM) algorithm (Dempster et al., 1977) to find maximum likelihood parameters $\boldsymbol{q}, \boldsymbol{\theta}$ of $X$. The EM algorithm begins with an arbitrary initial parameter assignment, $\boldsymbol{q}^0, \boldsymbol{\theta}^0$. It then repeats the two steps below, updating the parameter set, until convergence. After the $k$th iteration we start with parameters $\boldsymbol{q}^k, \boldsymbol{\theta}^k$:

**Expectation Step.** Using the current set of parameters, we define for each $\sigma[i] \in \mathcal{D}$, the probability density

$$r^k(h[i]) = p(h[i] \mid \sigma[i], \boldsymbol{q}^k, \boldsymbol{\theta}^k) \quad (1)$$

We then compute expected sufficient statistics $\bar{T}[x]$, $\bar{M}[x, x']$, and $\bar{M}[x]$ according to this posterior density over completions of the data given the data and the model.

**Maximization Step.** Using the expected sufficient statistics we just computed as if they came from a complete data set, we set $\boldsymbol{q}^{k+1}, \boldsymbol{\theta}^{k+1}$ to be the new maximum likelihood parameters for our model as follows

$$q_x^{k+1} = \frac{\bar{M}[x]}{\bar{T}[x]}; \quad \theta_{xx'}^{k+1} = \frac{\bar{M}[x,x']}{\bar{M}[x]}. \quad (2)$$

The difficult part in this algorithm is the Expectation Step. As we discussed, the space over which we are integrating is highly complex, and it is not clear how we can compute the expected sufficient statistics in a tractable way. This problem is the focus of the next section.

## 3 Computing Expected Sufficient Statistics

Given an $n$-state homogeneous Markov process $X_t$ with intensity matrix $\mathbf{Q}_X$, our task is to compute the expected sufficient statistics with respect to the posterior probability density over completions of the data given the observations and the current model. For simplicity, we omit the explicit dependence on the parameters $\boldsymbol{q}^k, \boldsymbol{\theta}^k$. Our analysis follows the work of Asmussen et al. (1996), who utilize numerical integration via the Runge-Kutta method. Holmes and Rubin (2002) provide an alternative approach using the eigenvalue decomposition of the intensity matrix.

### 3.1 Notation

In order to compute the expected sufficient statistics over $\mathcal{D}$, we compute them for each partially observed trajectory $\sigma \in \mathcal{D}$ separately and then combine the results.

A partially observed trajectory $\sigma$ is given as a sequence of $N$ subsystems so that the state is restricted to subsystem $S_i$ during interval $[t_i, t_{i+1})$ for $0 \leq i \leq (N-1)$. Without loss of generality, we assume that $\sigma$ begins at time 0 and ends at time $\tau$ so $t_0 = 0$ and $t_N = \tau$.

For subsystem $S$, let $\mathbf{Q}_S$ be the $n \times n$ intensity matrix $\mathbf{Q}_X$ with all intensities zeroed out except those corresponding to transitions within the subsystem $S$ (and associated diagonal elements). For subsystems $S_1, S_2$, let $\mathbf{Q}_{S_1 S_2}$ be the $n \times n$ intensity matrix $\mathbf{Q}_X$ with all intensities zeroed out except those corresponding to transitions from $S_1$ to $S_2$. Note that this means all the intensities corresponding to transitions within $S_1$ and within $S_2$ are also zeroed out.

Sometimes it is convenient to refer to evidence by providing an arbitrary time interval. Let $\sigma_{t_1:t_2}$ denote the evidence provided by $\sigma$ over the interval $[t_1, t_2)$. (So, $\sigma_{0:\tau}$ is the evidence provided by all of $\sigma$.) Let $\sigma_{t_1:t_2^+}$ denote the evidence over the interval $[t_1, t_2]$, and $\sigma_{t_1^-:t_2}$ the evidence over the interval $(t_1, t_2)$.

Let $\mathbf{e}$ be a (column) $n$-vector of ones. Let $\mathbf{e}_j$ be an $n$-vector of zeros with a one in position $j$. Let $\boldsymbol{\Delta}_{j,k}$ be an $n \times n$ matrix of zeros with a one in position $j, k$. (So $\boldsymbol{\Delta}_{j,k} = \mathbf{e}_j \mathbf{e}_k'$.) Note that all multiplications below are standard vector and matrix multiplications as opposed to factor multiplications.

Define the vectors $\boldsymbol{\alpha}_t^-$ and $\boldsymbol{\beta}_t^+$ component-wise as

$$\boldsymbol{\alpha}_t^-[i] = p(X_{t^-} = i, \sigma_{0:t})$$
$$\boldsymbol{\beta}_t^+[i] = p(\sigma_{t^+:\tau} \mid X_{t^+} = i),$$

where, if $X$ transitions at $t$, $X_{t^-}$ is the value of $X$ just prior to the transition, and $X_{t^+}$ the value just afterward. (If there is no transition, $X_{t^-} = X_{t^+}$.) Moreover, recall that $\sigma_{0:t}$ represents the evidence over interval $[0, t)$ not including $t$ and $\sigma_{t^+:\tau}$ represents evidence over interval $(t, \tau)$ not including $t$. Thus, neither of these vectors include evidence of a transition at time $t$. We also define the vectors

$$\boldsymbol{\alpha}_t[i] = p(X_t = i, \sigma_{0:t^+})$$
$$\boldsymbol{\beta}_t[i] = p(\sigma_{t:\tau} \mid X_t = i)$$

both of which include evidence of any transition at time $t$.

## 3.2 Expected Amount of Time

The sufficient statistic $T[j]$ is the amount of time that $X$ spends in state $j$ over the course of trajectory $\sigma$. We can write the expectation of $T[j]$ according to the posterior probability density given the evidence as

$$\mathbf{E}[T[j]] = \int_0^\tau p(X_t \mid \sigma_{0:\tau})\mathbf{e}_j dt$$
$$= \sum_{i=0}^{N-1} \int_{t_i}^{t_{i+1}} p(X_t \mid \sigma_{0:\tau})\mathbf{e}_j dt$$
$$= \frac{1}{p(\sigma_{0:\tau})} \sum_{i=0}^{N-1} \int_{t_i}^{t_{i+1}} p(X_t, \sigma_{0:\tau})\mathbf{e}_j dt .$$

The constant fraction at the beginning of the last line serves to make the total expected time over all $j$ sum to $\tau$.

We must show how to compute the above integrals over intervals of constant evidence. Let $[v, w)$ be such an interval, and let $S$ be the subsystem to which the state is restricted on this interval. Then we have

$$\int_v^w p(X_t, \sigma_{0:\tau})\mathbf{e}_j dt \qquad (3)$$
$$= \int_v^w p(X_t, \sigma_{0:t})\mathbf{\Delta}_{j,j} p(\sigma_{t:\tau} \mid X_t) dt$$
$$= \int_v^w \boldsymbol{\alpha}_v p(X_t, \sigma_{v:t} \mid X_v)\mathbf{\Delta}_{j,j} p(X_w, \sigma_{t:w} \mid X_t)\boldsymbol{\beta}_w dt$$
$$= \int_v^w \boldsymbol{\alpha}_v \exp(\mathbf{Q}_S(t-v))\mathbf{\Delta}_{j,j} \exp(\mathbf{Q}_S(w-t))\boldsymbol{\beta}_w dt .$$

The code accompanying Asmussen et al. (1996) is easily extendable to this more general case and can compute the above integral via the Runge-Kutta method of fourth order. This method traverses the interval in small discrete steps each of which has a constant number of matrix multiplications. Thus, the main factor in the complexity of this algorithm is the number of steps which is a function of the step size. Importantly, the integration uses a step size that is *adaptive* and not fixed. The intensities of the $\mathbf{Q}_S$ matrix represent rates of evolution for the variables in the cluster, so larger intensities mean a faster rate of change which usually requires a smaller step size. We begin with a step size proportional to the inverse of the largest intensity in $\mathbf{Q}_S$. The step size thus varies across different subsystems and is sensitive to the current evidence. Also, following Press et al. (1992), we use a standard adaptive procedure that allows larger steps to be taken when possible based on error estimates.

We can calculate the total expected time, $\mathbf{E}[T[j]]$, by summing the above expression over all intervals of constant evidence.

## 3.3 Expected Number of Transitions

The sufficient statistic $M[j, k]$ ($j \neq k$) is the number of times $X$ transitions from state $j$ to state $k$ over the course of trajectory $\sigma$. Following the derivation in Asmussen et al. (1996), we consider discrete time approximations of $M[j, k]$ and take the limit as the size of our discretization goes to zero, yielding an exact equation. For $\epsilon > 0$, let

$$M_\epsilon[j, k] = \sum_{t=0}^{\tau/\epsilon - 1} \mathbf{1}\{X_{t\epsilon} = j, X_{(t+1)\epsilon} = k\} .$$

Note that our discrete approximation is dominated by the actual value, i.e., $|M_\epsilon[j,k]| \leq M[j,k]$, and also that as $\epsilon \downarrow 0$, $M_\epsilon[j,k] \to M[j,k]$. Hence, by dominated convergence for conditional expectations, we have

$$\mathbf{E}[M[j,k]] = \lim_{\epsilon \downarrow 0} \mathbf{E}[M_\epsilon[j,k]] .$$

This last expectation can be broken down as

$$\mathbf{E}[M_\epsilon[j,k]] = \sum_{t=0}^{\tau/\epsilon - 1} \frac{p(X_{t\epsilon} = j, X_{(t+1)\epsilon} = k, \sigma_{0:\tau})}{p(\sigma_{0:\tau})} .$$

Note that, for $\epsilon$ small enough, we observe at most one transition per interval. Thus, each of the intervals in the sum falls into one of two categories: either the interval contains a (partially observed) transition, or the evidence is constant over the interval. We treat each of these cases separately.

Let $[t\epsilon, (t+1)\epsilon)$ be an interval containing a partially observed transition at time $t_i$. We observe only that we are transitioning from one of the states of $S_i$ to one of the states of $S_{i+1}$. We can calculate the contribution of this interval to the expected sufficient statistics (ignoring the constant $1/p(\sigma_{0:\tau})$) as

$$p(X_{t\epsilon} = j, X_{(t+1)\epsilon} = k, \sigma_{0:\tau})$$
$$= p(X_{t\epsilon} = j, \sigma_{0:t\epsilon})p(X_{(t+1)\epsilon} = k, \sigma_{t\epsilon:(t+1)\epsilon} \mid X_{t\epsilon} = j)$$
$$\quad p(\sigma_{(t+1)\epsilon:\tau} \mid X_{(t+1)\epsilon} = k) .$$

As $\epsilon \downarrow 0$, we have the probability of the state and the evidence up to, but not including, time $t_i$, times the instantaneous probability of transitioning from state $j$ to state $k$, times the probability of the evidence given the state just after $t_i$. Thus, at the limit, this transition's contribution is

$$\boldsymbol{\alpha}_{t_i}^- \mathbf{e}_j q_{jk} \mathbf{e}_k' \boldsymbol{\beta}_{t_i}^+ = q_{jk} \boldsymbol{\alpha}_{t_i}^- \mathbf{\Delta}_{j,k} \boldsymbol{\beta}_{t_i}^+ . \qquad (4)$$

Now consider the case when we are within an interval $[v, w) = [t_i, t_{i+1})$ of constant evidence — i.e., it does not contain a partially observed transition and will generally be of length much larger than $\epsilon$. Let $\Delta t = w - v$ and let $S$ be the subsystem to which the state is restricted on this interval. As $\epsilon$ grows small, the contribution of this interval

to the sum (again, ignoring $1/p(\sigma_{0:\tau})$) is

$$\sum_{t=0}^{\Delta t/\epsilon - 1} p(X_{v+t\epsilon} = j, X_{v+(t+1)\epsilon} = k, \sigma_{0:\tau})$$

$$= \sum_{t=0}^{\Delta t/\epsilon - 1} \sum_{X_v, X_w} p(X_v, \sigma_{0:v}) p(X_{v+t\epsilon} = j, \sigma_{v:v+t\epsilon} \mid X_v)$$
$$p(X_{v+(t+1)\epsilon} = k, \sigma_{v+t\epsilon:v+(t+1)\epsilon} \mid X_{v+t\epsilon} = j)$$
$$p(X_w, \sigma_{v+(t+1)\epsilon:w} \mid X_{v+(t+1)\epsilon} = k) p(\sigma_{w:\tau} \mid X_w)$$

$$= \sum_{t=0}^{\Delta t/\epsilon - 1} \boldsymbol{\alpha}_v \exp(\mathbf{Q}_S t\epsilon) \mathbf{e}_j$$
$$\mathbf{e}'_j p(X_{v+(t+1)\epsilon}, \sigma_{v+t\epsilon:v+(t+1)\epsilon} \mid X_{v+t\epsilon}) \mathbf{e}_k$$
$$\mathbf{e}'_k \exp(\mathbf{Q}_S(w - (v + (t+1)\epsilon))) \boldsymbol{\beta}_w .$$

As in the case with observed transitions, as $\epsilon \downarrow 0$, the middle term becomes $q_{jk}dt$, the instantaneous probability of transitioning. Since $\exp(\mathbf{Q}_S t)$ is continuous, we can express the limit as a sum of integrals of the form

$$q_{jk} \int_v^w \boldsymbol{\alpha}_v \exp(\mathbf{Q}_S(t-v)) \boldsymbol{\Delta}_{j,k} \exp(\mathbf{Q}_S(w-t)) \boldsymbol{\beta}_w dt \quad (5)$$

We have one such term for each interval of constant evidence. Essentially we are integrating the instantaneous probability of transitioning from state $j$ to $k$ over the interval given the evidence. Note that this is very similar to the form of Eq. (3) — the only difference is the matrix $\boldsymbol{\Delta}_{j,k}$ and the term $q_{jk}$.

To obtain the overall sufficient statistics, we have a sum with two types of terms: a term as in Eq. (4) for each observed transition, and an integral as in Eq. (5) for each interval of constant evidence. The overall expression is

$$\frac{q_{jk}}{p(\sigma_{0:\tau})} \left[ \sum_{i=1}^{N-1} \boldsymbol{\alpha}_{t_i}^- \boldsymbol{\Delta}_{j,k} \boldsymbol{\beta}_{t_i}^+ \right.$$
$$\left. + \sum_{i=0}^{N-1} \int_v^w \left( \begin{array}{c} \boldsymbol{\alpha}_v \exp(\mathbf{Q}_S(t-v)) \boldsymbol{\Delta}_{j,k} \\ \times \exp(\mathbf{Q}_S(w-t)) \boldsymbol{\beta}_w \end{array} \right) dt \right] .$$

### 3.4 Computing $\boldsymbol{\alpha}_t$ and $\boldsymbol{\beta}_t$

One method of computing $\boldsymbol{\alpha}_t$ and $\boldsymbol{\beta}_t$ is via a forward-backward style algorithm (Rabiner & Juang, 1986) over the entire trajectory to incorporate evidence and get distributions over the state of the system at every time $t_i$.

We have already defined the forward and backward probability vectors, $\boldsymbol{\alpha}_t$ and $\boldsymbol{\beta}_t$. To initialize the vectors, we simply let $\boldsymbol{\alpha}_0$ be the initial distribution over the state and $\boldsymbol{\beta}_\tau = \mathbf{e}$, a vector of ones. To update the vectors from their previously computed values, we calculate

$$\boldsymbol{\alpha}_{t_{i+1}} = \boldsymbol{\alpha}_{t_i} \exp(\mathbf{Q}_{S_i}(t_{i+1} - t_i)) \mathbf{Q}_{S_i S_{i+1}}$$
$$\boldsymbol{\beta}_{t_i} = \mathbf{Q}_{S_{i-1} S_i} \exp(\mathbf{Q}_{S_i}(t_{i+1} - t_i)) \boldsymbol{\beta}_{t_{i+1}}$$

To exclude incorporation of the evidence of the transition from either forward or backward vector (or if the time in question is not a transition time), one can simply remove the subsystem transition intensity matrix ($\mathbf{Q}_{S_i S_{i+1}}$) from the calculation. For example, as time 0 and $\tau$ are not transition times, we have

$$\boldsymbol{\alpha}_\tau = \boldsymbol{\alpha}_{t_{N-1}} \exp(\mathbf{Q}_{S_{N-1}}(\tau - t_{N-1}))$$
$$\boldsymbol{\beta}_0 = \exp(\mathbf{Q}_{S_0}(t_1 - 0)) \boldsymbol{\beta}_{t_1} .$$

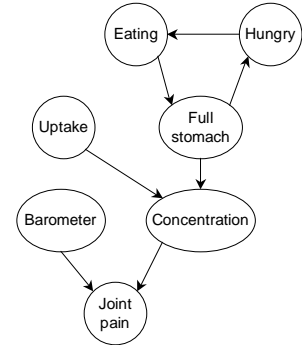

Figure 1: Drug effect network

We can also compute

$$\boldsymbol{\alpha}_{t_{i+1}}^- = \boldsymbol{\alpha}_{t_i} \exp(\mathbf{Q}_{S_i}(t_{i+1} - t_i))$$
$$\boldsymbol{\beta}_{t_i}^+ = \exp(\mathbf{Q}_{S_i}(t_{i+1} - t_i)) \boldsymbol{\beta}_{t_{i+1}} .$$

We can then write the distribution over the state of the system at time $t$ given all the evidence as

$$P(X_t = j \mid \sigma_{0:\tau}) = \frac{1}{p(\sigma_{0:\tau})} \boldsymbol{\alpha}_t^- \boldsymbol{\Delta}_{j,j} \boldsymbol{\beta}_t .$$

## 4 CTBNs

We can now extend the EM algorithm to continuous time Bayesian networks which are a factored representation for homogeneous Markov processes.

### 4.1 Continuous Time Bayesian Networks

We briefly review continuous time Bayesian networks as presented in Nodelman et al. (2003). A CTBN represents a stochastic process over a structured state space, consisting of assignments to some set of local variables $\boldsymbol{X} = \{X_1, X_2, \ldots, X_k\}$.

We model the joint dynamics of these local variables by allowing the transition model of each local variable to be a Markov process whose parameterization depends on some subset of other variables $\mathbf{U}$. The key building block is a *conditional Markov process*:

**Definition 4.1** *A* conditional Markov process *$X$ is an inhomogeneous Markov process whose intensity matrix varies with time, but only as a function of the current values of a set of discrete conditioning variables $\mathbf{U}$. Its intensity matrix, called a* conditional intensity matrix *(CIM), is* written $\mathbf{Q}_{X|\mathbf{U}}$ and can be viewed as a set of homogeneous intensity matrices $\mathbf{Q}_{X|\mathbf{u}}$ — one for each instantiation of values $\mathbf{u}$ to $\mathbf{U}$. ∎

The parameters of $\mathbf{Q}_{X|\mathbf{U}}$ are $\boldsymbol{q}_{X|\mathbf{u}} = \{q_{x|\mathbf{u}} : x \in \textit{Val}(X)\}$ and $\boldsymbol{\theta}_{X|\mathbf{u}} = \{\theta_{xx'|\mathbf{u}} : x, x' \in \textit{Val}(X), x \neq x'\}$.

We can now combine a set of conditional Markov processes to form a CTBN:

**Definition 4.2** *A continuous time Bayesian network $\mathcal{N}$ over $\boldsymbol{X}$ consists of two components: an initial distribution $\boldsymbol{P}_{\boldsymbol{X}}^0$, specified as a Bayesian network $\mathcal{B}$ over $\boldsymbol{X}$, and a continuous transition model, specified as*

- *A directed (possibly cyclic) graph $\mathcal{G}$ whose nodes are $X_1, \ldots, X_k$; $\boldsymbol{Pa}_{\mathcal{G}}(X_i)$, often abbreviated $\mathbf{U}_i$, denotes the parents of $X_i$ in $\mathcal{G}$.*
- *A conditional intensity matrix, $\mathbf{Q}_{X_i|\mathbf{U}_i}$, for each variable $X_i \in \boldsymbol{X}$.* ∎

**Example 4.3** *Figure 1 shows the graph structure for a modified version of the drug effect CTBN network from Nodelman et al. (2002) modelling the effect of a drug a person might take to alleviate pain in their joints. There are nodes for the uptake of the drug and for the resulting concentration of the drug in the bloodstream. The concentration is affected by how full the patient's stomach is. The pain may be aggravated by changing pressure. The model contains a cycle, indicating that whether a person is hungry depends on how full their stomach is, which depends on whether or not they are eating, which in turn depends on whether they are hungry.*

### 4.2 Expected Log-likelihood

Extending the EM algorithm to CTBNs involves making it sensitive to a factored state space. Our incomplete data, $\mathcal{D}$, are now partially observed trajectories describing the behavior of a dynamic system factored into a set of state variables $\boldsymbol{X}$.

As shown by Nodelman et al. (2003), the log-likelihood decomposes as a sum of local log-likelihoods for each variable. Specifically, given variable $X$, let $\mathbf{U}$ be its parent set in $\mathcal{N}$. Then the sufficient statistics of $\mathcal{D}^+$ for our model are $T[x|\mathbf{u}]$, the total amount of time that $X = x$ while $\mathbf{U} = \mathbf{u}$, and $M[x, x'|\mathbf{u}]$, the number of times $X$ transitions from $x$ to $x'$ while $\mathbf{U} = \mathbf{u}$. If we let $M[x|\mathbf{u}] = \sum_{x'} M[x, x'|\mathbf{u}]$ the likelihood for each variable $X$ further decomposes as

$$\ell_X(\boldsymbol{q}, \boldsymbol{\theta} : \mathcal{D}^+) = \ell_X(\boldsymbol{q} : \mathcal{D}^+) + \ell_X(\boldsymbol{\theta} : \mathcal{D}^+)$$
$$= \left[\sum_{\mathbf{u}}\sum_x M[x|\mathbf{u}] \ln(q_{x|\mathbf{u}}) - q_{x|\mathbf{u}} \cdot T[x|\mathbf{u}]\right]$$
$$+ \left[\sum_{\mathbf{u}}\sum_x \sum_{x' \neq x} M[x,x'|\mathbf{u}] \ln(\theta_{xx'|\mathbf{u}})\right] . \quad (6)$$

By linearity of expectation, the expected log-likelihood function also decomposes in the say way, and we can write the expected log-likelihood $\mathbf{E}_r[\ell(\boldsymbol{q}, \boldsymbol{\theta} : \mathcal{D}^+)]$ as a sum of terms (one for each variable $X$) in the same form as Eq. (6), except using the expected sufficient statistics $\bar{T}[x|\mathbf{u}]$, $\bar{M}[x, x'|\mathbf{u}]$, and $\bar{M}[x|\mathbf{u}]$.

### 4.3 EM for CTBNs

The EM algorithm for CTBNs is essentially the same as for homogeneous Markov processes. We need only specify how evidence in the CTBN induces evidence on the induced Markov process, and how expected sufficient statistics in the Markov process give us the necessary sufficient statistics for the CTBN.

A CTBN is a homogeneous Markov process over the joint state space of its constituent variables. Any assignment of values to a subset of the variables forms a subsystem of the CTBN — it restricts us to a subset of the joint state space (as shown with binary "variables" $Y$ and $Z$ in Example 2.1). Just as before, our evidence can be described as a sequence of subsystems $S_i$, each with an associated duration.

Recall that, in a CTBN, the expected sufficient statistics have the form $\bar{T}[x|\mathbf{u}]$ and $\bar{M}[x, x'|\mathbf{u}]$. We can thus replace Eq. (2) in the maximization step of EM with:

$$q_{x|\mathbf{u}}^{k+1} = \frac{\bar{M}[x|\mathbf{u}]}{\bar{T}[x|\mathbf{u}]}; \quad \theta_{xx'|\mathbf{u}}^{k+1} = \frac{\bar{M}[x,x'|\mathbf{u}]}{\bar{M}[x|\mathbf{u}]} . \quad (7)$$

The expectation step of EM can, in principle, be done by flattening the CTBN into a single homogeneous Markov process with a state space exponential in the number of variables and following the method described above. In this case, we can compute $\bar{T}[x|\mathbf{u}]$ by summing up all of the expected sufficient statistics $\bar{T}[j]$ for any state $j$ consistent with $X = x, \mathbf{U} = \mathbf{u}$. Similarly, $\bar{M}[x|\mathbf{u}]$ can be computed by summing up all of $\bar{M}[j, k]$ for state $j$ consistent with $X = x, \mathbf{U} = \mathbf{u}$ and $k$ consistent with $X = x', \mathbf{U} = \mathbf{u}$.

However, as the number of variables in the CTBN grows, that process becomes intractable, so we are forced to use approximate inference. The approximate inference algorithm must be able to compute approximate versions of the forward and backward messages $\boldsymbol{\alpha}_t, \boldsymbol{\beta}_w$. It must also be able to extract the relevant sufficient statistics — themselves a sum over an exponentially large space — from the approximate messages efficiently.

A companion paper (Nodelman et al., 2005) provides a cluster graph inference algorithm which can be used to perform this type of approximate inference. For each segment $[t_i, t_{i+1})$ of continuous fixed evidence, we construct a cluster graph data structure, whose nodes correspond to clusters of variables $\mathcal{C}_k$, each encoding a distribution over the trajectories of the variables $\mathcal{C}_k$ for the duration $[t_i, t_{i+1})$. A message-passing process calibrates the clusters. We can then extract from the cluster $\mathcal{C}_k$ both beliefs about the momentary state of the variables $\mathcal{C}_k$ at time $t_i$ and $t_{i+1}$, as well as a distribution over the trajectories of $\mathcal{C}_k$ during the interval. The former provide a factored representation of our forward message $\boldsymbol{\alpha}_{t_{i+1}}$ and backward message $\boldsymbol{\beta}_{t_i}$, and are incorporated into the cluster graphs for the adjoining cluster in a forward-backward message passing process. The cluster distributions are represented as local intensity matrices, from which we can compute the expected sufficient statistics over families $X_i, \mathbf{U}_i$, as above. This, this algorithm allows us to perform the steps required for the E-step, and the M-step can be performed easily as described above.

### 4.4 Structural EM for CTBNs

We can also learn structure from incomplete data by applying the structural EM (SEM) algorithm of Friedman (1997) to our setting. We start with some initial graph structure $\mathcal{G}^0$, initial parameters $\boldsymbol{q}^0, \boldsymbol{\theta}^0$, and dataset $\mathcal{D}$.

At each iteration, as with SEM for Bayesian networks, we choose between taking a step to update the parameters and taking a step to modify the structure.

**Parameter Update Step.** Run the EM algorithm, computing new expected sufficient statistics and updating the parameters as in Eq. (7).

**Structure Modification Step.** Using the current parameterization and expected sufficient statistics, choose a structure modification that increases the score. SEM is used with a variety of scores, most commonly the BIC score or the Bayesian score with expected sufficient statistics as if real. In both cases, the score can be written in terms of expected sufficient statistics, allowing SEM to be used with our algorithm above.

SEM leaves unspecified the issue of how many greedy search steps one takes before recomputing the expected sufficient statistics and parameters. Nodelman et al. (2003) showed that, for CTBNs, structure search for a fixed number of parents per node can be done in polynomial time. Thus, it is possible, in this setting, to find the globally optimal structure given the current parametrization in the structure modification step. If one does this, SEM for CTBNs becomes an iterated optimization algorithm with a full maximization step for both structure and parameters.

## 5 Complex Duration Distributions

An important limitation to the expressive power in continuous time Bayesian networks has been the restriction to modelling durations in a single state as exponential distributions over time. With the extension of the EM to CTBNs, we can now address this limitation.

### 5.1 Phase Distributions

*Phase distributions* are a rich, semi-parametric class of distributions over durations, that use the exponential distribution as a building blcok. A phase distribution is modeled as a set of *phases*, through which a process evolves. Each of these phases is associated with an exponential distribution, which encodes the duration that a process stays in that phase. That is, we enter a phase $k$, and then leave in time $t$ exponentially distributed with the parameter $q_k$ associated with that phase. We can view the process as moving over a directed, possibly cyclic graph, consisting of these phases. Thus, we can create combinations of chains, mixtures, and loops of such exponentially distributed phases linked together in a variety of ways. We spend some amount of time going from one phase to another, but eventually we leave the set of phases altogether. The distribution over when we leave such a system of phases is called a *phase distribution*.

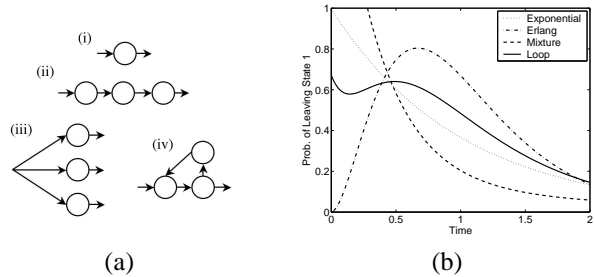

Figure 2: (a): Phase transition diagrams for (i) a single exponential phase, (ii) an Erlang (chain), (iii) a mixture, and (iv) a loop. (b): The probability density of the first time to leave state 1 (the phase distribution), for three example binary variable with 3 phases for both states. All examples have an expected time of 1 to leave state 1.

**Example 5.1** *Consider a 4-state homogeneous Markov process $PH_t$ with intensity matrix*

$$\mathbf{Q}_{PH} = \begin{bmatrix} -q_1 & q_{12} & q_{13} & q_{14} \\ q_{21} & -q_2 & q_{23} & q_{24} \\ q_{31} & q_{32} & -q_3 & q_{34} \\ 0 & 0 & 0 & 0 \end{bmatrix}.$$

*If the intensities of states 1, 2, and 3 are non-zero, then regardless of the initial phase, $PH_t$ will end up in state 4 and remain there. Thus, state 4 is called an* absorbing *state and the others are called* transient *states. We call the transient states* phases *and the distribution over when $PH_t$ reaches state 4 is called a* phase *distribution. In this particular case it has 3 phases. If we wanted to encode a chain $1 \to 2 \to 3$, we would have all off-diagonal entries equal to 0, except $q_{12}, q_{23}, q_{34}$. If we wanted to encode a loop, we would also allow $q_{31} \neq 0$. Figure 2 shows some simple distribution shapes that can be formed with 3 phases. Note that, while a chain distribution always begin in phase 1 and ends in some final phase $p$, general phase distributions might start and end in any phase (e.g., the mixture distribution shown).*

**Definition 5.2** *A phase distribution of $p$ phases is defined as the distribution over time when a homogeneous Markov process with a single absorbing state and $p$ transient phases reaches absorption (Neuts 1975; 1981).*

We can specify a $p$-phase distribution with a $p \times p$ matrix, $\mathbf{Q}_P$, by including only the subsystem of transient phases without losing any information. The rows of the new intensity matrix will have a (possibly) negative row sum, where the missing intensity corresponds exactly to the intensity with which we leave the entire system.

Phase distributions with a single phase are simply exponential distributions. The general class is highly expressive: any distribution can be approximated with arbitrary precision by a phase distribution with some finite number of phases (Neuts, 1981). A commonly used subclass of phase distributions is the *Erlangian-p* which can be constructed with a chain-structured subsystem of $p$ phases, where all phases have the same exit intensity.

## 5.2 CTBN Durations as Phase Distributions

One can directly model the distribution in state $x$ of variable $X$ in CTBN $\mathcal{N}$ as any *phase distribution* instead of an exponential distribution. This idea is first described by Nodelman and Horvitz (2003) and subsequently in Gopalratnam et al. (2005). The former focuses on Erlang distributions and the latter on Erlang-Coxian distributions both of which are limited subclasses of general phase distributions. In particular neither allow for the exponential phases to be looped. Restriction to subclasses makes learning from data easier; in particular, EM is not required necessary to learn distributions in these classes; however, these subclasses have several drawbacks, including reduced expressivity, especially with small numbers of phases (Asmussen et al., 1996). Our method, as it is based on a general EM algorithm, allows the use of general phase distributions in CTBNs without restriction.

When using this *phase modelling* method, the structure of the intensity matrix must be altered by adding *phases* as additional rows (and columns). We use the term *phases* to distinguish additional hidden state in the intensity matrix from states of the variable. Thus, a subsystem of several phases is used to implement a single state of a variable. In this context there is no absorption and the "final" transition of the phase distribution is the transition of the variable to its the next state.

**Example 5.3** *To make a binary variable W have the duration in each of its 2 states be Erlangian-3 distributions (with distinct parameters), we write its intensity matrix as*

$$\mathbf{Q}_W = \begin{bmatrix} -1 & 1 & 0 & 0 & 0 & 0 \\ 0 & -1 & 1 & 0 & 0 & 0 \\ 0 & 0 & -1 & 1 & 0 & 0 \\ 0 & 0 & 0 & -2 & 2 & 0 \\ 0 & 0 & 0 & 0 & -2 & 2 \\ 2 & 0 & 0 & 0 & 0 & -2 \end{bmatrix}.$$

*The top three rows correspond to state $w_1$ and the bottom three to state $w_2$. Note that when restricted to modelling with Erlang distributions for a fixed number of phases, the number of free parameters is the same as a regular (exponentially distributed) CTBN.*

Using *phase modelling* greatly extends the expressive power of CTBNs and fits naturally within the existing CTBN framework. The basic structure of existing algorithms for CTBNs remains unaltered.

The child of a variable with complex durations sees only the state of its parent and so does not, in general, depend on the current phase of a parent. There are a number of design choices to make in implementing phase distributions for durations in CTBNs. Different choices may be appropriate for different applications. One can add phases in a uniform way to each state of each variable — as in example 5.3 where each state gets three phases. Alternatively, one might allow some states of some variables to be modelled with more phases than others.

When the parent instantiation changes, one might allow the child in its current state to stay in the same phase or to reset. Care must be taken when allowing the child to stay in the same phase — it requires consistency at least in the number of phases allowed for each state across all parent instantiations. When the phase distribution does not have a distinguished start phase, there is also a choice about the distribution over phases with which one enters a state. In particular, there might be a fixed distribution over phases with which one always enters a particular state or that distribution might depend upon the previous state or the current parent instantiation.

Our observations of a phase-distributed variable are always partial in that we might observe the current *state* of a variable but never its associated *phase*. We can learn the parameters for a CTBN with phase distributions using the EM algorithm described above by viewing it as a regular partially observed process.

## 5.3 Using Hidden Variables

An alternate method of allowing complex duration distributions for the states of a variable $X$ in CTBN $\mathcal{N}$ is to introduce a special hidden variable $H_X$ as a parent of $X$. This has the advantage of being a very clean way to add expressive power to CTBNs. Without this technique, the parents of $X$ must be other variables that we are modelling in our domain which means that the intensities which control the evolution of $X$ can only change when a regular modelled variable changes. The addition of a hidden parent allows us to describe more complex distributions over trajectories of $X$ by allowing the intensities which control the evolution of $X$ to change more frequently.

There are different ways to add a hidden variable $H_X$ as a parent of $X$. We might force $H_X$ to have no parents, or allow it to have parents in addition to $X$ having parents. However, while adding an explicit hidden variable with clear semantics might be useful, it is important to realize that all complex duration distributions expressible through use of hidden variables are expressible by direct phase modelling of the state. For example, suppose $H_X$ has 3 states and $X$ has 2 states. Using direct phases, we can rewrite a $6 \times 6$ intensity matrix for $X$ having 3 phases for each of its 2 states corresponding to the hidden state of $H_X$. More generally, we can amalgamate a set of hidden parents $\boldsymbol{H}_X$ and $X$ into a single cluster node $\boldsymbol{S}$, whose

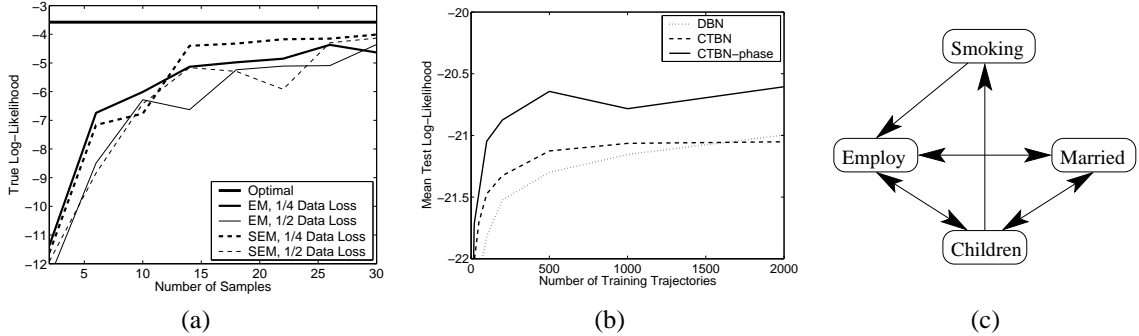

Figure 3: (a) Learning results for drug effect net. (b) Learning results for British Household Panel Survey. (c) Learned BHPS network (200 training points).

parents are the union of $X$'s parents other than $\boldsymbol{H}_X$ and the parents of $\boldsymbol{X}_X$. Each state of $X$ now corresponds to a set of instantiations to $\boldsymbol{S}$. We can reinterpret the amalgamated CIM for $\boldsymbol{S}$ (given its parents) as a phase distribution for $X$, with $|Val(\boldsymbol{H}_X)|$ phases per state of $X$. Thus, we can show:

**Theorem 5.4** *For a fixed number of phases $p$, a CTBN variable $X$ with direct phase modelling for durations, i.e., using $|X| \cdot p$ rows for $\mathbf{Q}_X$, is strictly more expressive than a variable with complex durations modelled by using a hidden parent $H_X$ with $p$ states.*

Conversely, many complex duration distributions expressible by direct phase modelling cannot be expressed using hidden variables. In particular, the joint behavior of $H_X$ and $X$ is restricted in that $X$ and $H_X$ cannot transition simultaneously. This corresponds to the constraint that $\mathbf{Q}_S$ must have zeros in locations that correspond to a simultaneous shift in the state of $X$ and $H_X$. No such constraint holds in direct phase modelling which means that we have more free parameters to describe the distribution.

## 6 Results

We implemented the EM and SEM algorithms described above. We used exact inference by constructing the flattened state space. While this prevents us from solving large problems, it also keeps the analysis of EM separate from that of approximate inference. To verify our algorithms' correctness, we used the drug effect network from Example 4.3, where all of the variables were binary-valued, for tractability. We sampled increasing numbers of trajectories of 5 time lengths. We ran both EM and SEM, giving the former the true network structure and hiding the structure from the latter. For each data example, we hid parts of the trajectory by selecting time windows of length 0.25 uniformly at random and hiding the value of one variable during the window. We kept dropping data in this fashion until all variables had lost either 1/4 or 1/2 of their total trajectory, depending on the experiment. The results of these experiments are shown in Figure 3(a). In some cases SEM outperforms EM because the true structure with learned parameters yields a lower log-likelihood given the amount of data at those points. Note that the horizontal axis represents the amount of data prior to dropping any of it.

The SEM algorithm worked well in this setting. As noted, removing the restriction on acyclicity allows CTBN structure search to decompose. Therefore, at each iteration, we performed a full structure search, which provided marked improvement over a greedy one-step optimization.

We also ran SEM on the British Household Panel Survey (ESRC Research Centre on Micro-social Change, 2003). This data is collectedyearly asking thousands of residents of Britain about important events in their lives.We randomly divided this set into 4000 training examples and 4000 testing examples (each example is a trajectory of a different person). Because we are employing exact inference, we had to keep the variable set small and chose 4 variables: employ (ternary: student, employed, unemployed), married (binary: not married, married), children (ternary: 0, 1, 2+), and smoking (binary: non-smoker, smoker). The average number of events per person is 5.6.

We learned structures and parameters for a time-sliced DBN with a time-slice width of 1 year, a standard CTBN, and a CTBN with 2 phases for every state of every variable. No restrictions were placed on the structure of these 2 phases (so, in general, they form a loop). In order to compare CTBNs to DBNs, we sampled the testing data at the same yearly rate and calculated the probability of these sampled trajectories.The results are shown in Figure 3(b). The DBN and plain CTBN models are comparable, with the DBN doing better with more data due to its increased flexibility (due to intra-time slice arcs, DBNs have more potential parameters). However, the phase distributions increase the performance of the CTBN model; the trajectories are approximately twice as likely as with the other two models. Figure 3(c) shows a learned exponential CTBN network. The parameters are interesting. For example, the rate (intensity) with which a person stops smoking given that they have two or more children is three times the rate at which a childless person quits smoking. The rate at which a person begins smoking given that they have no children is 300 times the rate of a person with two or more children. The rate at which a person becomes unemployed (after having been employed) tends to decrease with more children (unless the person does not smoke and is not married). The rate of becoming unemployed also tends to be less if one smokes (unless one is married and has a child).

## 7 Discussion

In this paper, we provided an algorithm for learning both structure and parameters of CTBNs from partially observed data. Given the scarcity of fully-observed data, particularly in the continuous-time setting, we believe that this development is likely to greatly increase the applicability of the CTBN framework to real-world problems.

Our experimental results were limited due to the reliance on exact inference. To scale up to larger problems, approximate inference algorithms must be employed. Our companion paper (Nodelman et al., 2005) pursues this topic.

This paper addresses one of the primary limitations of both CTBN and DBN models. Both models essentially assume an exponential (or geometric) model for state transitions. Although reasonable in some cases, an exponential model is a poor fit for many real-life domains (such as the interval between getting married and having children). Indeed, our experimental results show that CTBNs parameterized with the richer class of phase distributions significantly outperform both CTBNs and DBNs on a real-world domain. An important extension is to investigate the applicability of different phase distribution transition models, and to construct efficient algorithms for selecting an appropriate transition model automatically.

More globally, it would be interesting to apply phase-distribution CTBNs in other domains, and see whether this rich class of continuous-time models achieves improved performance over discrete-time models more broadly. In particular, many systems do not have a single natural time granularity; for example, in traffic modeling, locations of vehicles evolve more rapidly than driver intention, which evolves more quickly than the weather. Although one can learn discrete-time models that use the finest level of granularity, the geometric duration distribution of such models (which is particularly marked in fine-grained time models) can be a poor fit to the true duration distribution. Conversely, using an overly coarse granularity can also lead to artifacts in the learned model (Nodelman et al., 2003). We believe that the modeling flexibility resulting from continuous time models, augmented with the efficient structure learning made possible in CTBNs, will allow us to better tackle multiple applications, ranging from modeling people's activity to learning evolutionary models of DNA sequences.

**Acknowledgments.** This work was funded by DARPA's EPCA program, under subcontract to SRI International, and by the Boeing Corporation.